\documentclass{article}

\usepackage{microtype}
\usepackage{graphicx} % remove demo option once figures are there
\usepackage{booktabs} % for professional tables
\usepackage{mathtools} % amsmath with fixes and additions
\usepackage{siunitx} % for proper typesetting of numbers and units
\usepackage{booktabs} % commands to create good-looking tables
\usepackage{tikz} % nice language for creating drawings and diagrams
\usepackage{bm}

\usepackage{todonotes}

\usepackage{multirow}
\usepackage{paralist}

\usepackage{diagbox}
\usepackage{float}
\usepackage{subcaption}

\usepackage[ruled, linesnumbered, algo2e]{algorithm2e}

\newcommand{\B}[1]{\mathbf{#1}}

\usepackage{xurl}

\usepackage{hyperref}

\usepackage[accepted]{icml2021}

\icmltitlerunning{Monte Carlo EM for Deep Time Series Anomaly Detection}

\begin{document}

\twocolumn[
\icmltitle{Monte Carlo EM for Deep Time Series Anomaly Detection}



\icmlsetsymbol{equal}{*}

\begin{icmlauthorlist}
\icmlauthor{Fran\c{c}ois-Xavier Aubet}{aws}
\icmlauthor{Daniel Z\"{u}gner}{tum}
\icmlauthor{Jan Gasthaus}{aws}
\end{icmlauthorlist}

\icmlaffiliation{aws}{ AWS AI Labs}
\icmlaffiliation{tum}{Technical University of Munich}

\icmlcorrespondingauthor{Fran\c{c}ois-Xavier Aubet}{aubetf@amazon.com}

\icmlkeywords{Time Series, Anomaly Detection, Bayesian Inference}

\vskip 0.3in
]



\printAffiliationsAndNotice{}  % leave blank if no need to mention equal contribution

\begin{abstract}
	Time series data are often corrupted by outliers or other kinds of anomalies. Identifying 
	the anomalous points can be a goal on its own (anomaly detection), or a means to improving
	performance of other time series tasks (e.g.\ forecasting).
	Recent deep-learning-based approaches to anomaly detection and forecasting commonly assume that
	the proportion of anomalies in the training data is small enough to ignore, and treat the 
	unlabeled data as coming from the nominal data distribution.
	We present a simple yet effective technique for augmenting existing time series models so that they explicitly account for anomalies in the training data.
	By augmenting the training data with a latent anomaly indicator variable whose distribution is inferred while training the underlying model using Monte Carlo EM, our method simultaneously infers anomalous points while improving model performance on nominal data.
	We demonstrate the effectiveness of the approach by combining it with a simple feed-forward forecasting model.
	We investigate how anomalies in the train set affect the training of forecasting models, which are commonly used for time series anomaly detection, and show that our method improves the training of the model.

% When training a model to perform anomaly detection one has most often only access to an unlabeled dataset where most of the points are normal, but where there may be some anomalies.
% %
% It is very costly to label these anomalies, especially in time series anomaly detection.
% %
% Most method assume that the few anomalies will not affect the training of the model, without necessarily demonstrating or proving this.
% %
% We propose to infer at training time which points of the training set are anomalous and to replace them with their latent normal equivalent.
% %
% Our method can be used on top of any state of the art time series modeling model or time series anomaly detection model.
% %
% We investigate how anomalies in the train set affect the training of forecasting models, which are commonly used for time series anomaly detection, and show that our method improves the training of the model.
\end{abstract}

\section{Introduction}

%-> present the state of the art and the problem

In many time series anomaly detection applications one only has access to unlabeled data. This data is usually mostly nominal but may contain some (unlabeled) anomalies. 
Examples of this setting are e.g.\ the widely used anomaly detection benchmarks SMAP, MSL \citep{Hundman2018telemanon}, %SWaT \citep{mathur2016swat},
and SMD \citep{omnianomaly}.%, or the Yahoo dataset where 

This ``true'' unsupervised setting with \emph{mixed} data can be contrasted with the ``nominal-only'' setting, where one assumes access to ``clean'' nominal data.
%but where the anomalies are not labeled because of the cost of labeling anomalies. (cite SMAP, MSL, SWaT, SMD)
%
In practice, techniques that (explicitly or implicitly) assume access to nominal data can often also successfully be applied to mixed data by assuming it is nominal, as long as the proportion of anomalies is sufficiently small., they are however biased by training on some anomalous data.

While some time series anomaly detection model rely on the one class classification paradigm which does not suffer from this assumption \cite{thoc, ncad},
the vast majority of the current time series anomaly detection methods are either forecasting methods \citep{shipmon2017time, zhao2020multivariate} or reconstruction methods \citep{omnianomaly, donut, lstmvae, Zhang2019mscred}. Forecasting methods detect anomalies as deviations of observations from predictions, while reconstruction methods declare observations that deviate from the reconstruction as anomalous. 
%which aim at learning a model of $\B{y}^+$, and use it on incoming time points: if the new time point has a low probability under their model it is classified as being anomalous and coming from $\B{y}^-$. T
In both cases, a probabilistic model of the observed data is assumed and its parameters are learned.
However, by training the model on the observed data which contains both normal and anomalous data points, the model ultimately learns the wrong data distribution.
\citet{ehrlich2021spliced} propose an approach to make the model robust to the anomalous points, still the aim is to learn the distribution of both the normal and the anomalous points.
We propose to address this issue using a simple technique based on latent indicator variables that can readily be combined with existing probabilistic anomaly detection approaches. By using latent indicator variables to explicitly infer which observations in the training set are anomalous, we can subsequently suitably account for the anomalous observations while training the probabilistic model.
%

%Many recently proposed ``unsupervised'' anomaly detection techniques are actually of the latter type, 
%While some method propose to incorporate the anomalies of the training set in the training transforming it to a semi-supervised problem (cite NCAD), the vast majority of the approaches assumes that there are no anomalies and rely on the hope that the small number of anomalies will not affect the training of the model. (cite omnyanomaly and others)
%
Probabilistic models that use latent (unobserved) indicator variables to explicitly distinguish between nominal and anomalous data points are well-established in the context of robust mixture models \citep[e.g.][]{fraley1998many} and classical time series models \citep[e.g.][]{wang2018robust}. However, these techniques have not yet been utilized in the context of recent advances in \emph{deep} anomaly detection and time series modeling, presumably due to the (perceived) increased complexity of the required probabilistic inference and training procedure. We show that combining latent anomaly indicators with a Monte Carlo Expectation-Maximization (EM) \citep{wei1990monte} training procedure, results in a simple yet effective technique that can be combined with (almost) all existing deep anomaly detection and time series forecasting techniques.

%We propose a method to infer which of the points in the training set are anomalous and to handle them differently.

%We use a Hidden Markov Model (HMM) \cite{rabiner1986introduction} to model the latent time series of states, normal or anomalous. We use an Expectation Maximization (EM) \cite{dempster1977maximum} like procedure: in the E step we infer the state probabilities for each time point, in the M step we maximize the likelihood of the latent time series under the trained model, and we update the HMM state transition matrix. This model can be any of the forecasting or reconstruction based model mentioned above.

%While it is very rare to have access to reliable labels in time series anomaly detection datasets, it is possible for there to be some kind of prior knowledge on the portion of the points that could be anomalous, or a belief on how long anomalies last for: rather single outliers or long anomalous bands.
%or some noisy labels (often much bigger windows than the actual anomalies).
%Our method allows to encode these in the initialisation of the HMM.
%We may have access to a prior on the percentage of the points in the dataset that could be anomalous and some labels. We want to be able to take advantage of these, but not have to rely on it.

We demonstrate the effectiveness of our approach with a simple model for anomaly detection on the Yahoo anomaly detection dataset and on the electricity dataset for forecasting from a noisy training set.
%
%We make our code available for any one to use it to wrap their favourite time series modeling model\footnote{Link to github repository}.

\section{Background}
For non-time series data, one common approach of formalizing the notion of anomalies is to assume that the observed data is generated by a mixture model \citep{ruff2020unifying}: each observation $\B{x}$ is drawn from the mixture distribution $p(\B{x}) = \alpha p^{+}(\B{x}) + (1-\alpha)p^{-}(\B{x})$, where $p^{+}(\B{x})$ is the distribution of the nominal data and $p^{-}(\B{x})$ the anomalous data distribution. Typically one assumes a flexible parametrized distribution for $p^+$ and a broad, unspecific distribution for $p^{-}$ (e.g.\ a uniform distribution over the extent of the data).

This mixture distribution can equivalently be written using a binary \emph{indicator latent variable} $z$ taking value $0$ with probability $p(z=0) = \alpha$ and value 1 with probability $p(z=1)=1-\alpha$, and specifying the conditional distribution
\begin{equation}
	p(\B{x}|z) = \begin{cases}
		p^{+}(\B{x}) & \text{if}\ z = 0\\
		p^{-}(\B{x}) & \text{if}\ z = 1, 
	\end{cases}
\end{equation}
so that $p(\B{x}) = \sum_z p(\B{x}|z)p(z) = \alpha p^{+}(\B{x}) + (1-\alpha)p^{-}(\B{x})$. In this setup, anomaly detection can be performed by inferring the posterior distribution $p(z|\B{x})$ (and thresholding it if a hard choice is desired). Yet another way of representing the same model is generatively: first, draw $\B{y}^+ \sim p^+(\cdot)$, $\B{y}^- \sim p^-(\cdot)$, and $z \sim \text{Bernoulli}(1-\alpha)$, and then set $\B{x} = \mathbf{I}[z=0] \, \B{y}^+ + \mathbf{I}[z=1] \, \B{y}^-$, i.e.\ the observation $\B{x}$ is equal to $\B{y}^+$ if it is nominal ($z=1$) and equal to $\B{y}^-$ otherwise. Introducing the additional latent variables $\B{y}^+$ and $\B{y}^-$ is unnecessary in the IID setting, but becomes useful in the time series setting described next.

In time series setting, where the the observations are time series $\B{x}_{1:T} = \B{x}_1, \ldots, \B{x}_T$ that exhibit temporal dependencies, and anomalies are time points or regions within these time series, we have one anomaly indicator variable $z_t$ corresponding to each time point $\B{x}_t$. Like before, the nominal data is drawn from a parametrized probabilistic model $p_\theta^+(\B{y}_{1:T})$, and the anomalies are generated from a fixed model $p^-(\B{y}_{1:T})$. For time series data, the mixture data model then amounts to drawing $\B{y}_{1:T}^+ \sim p^+(\cdot)$, $\B{y}^-_{1:T} \sim p^-(\cdot)$, and $z_{1:T} \sim p^z(z_{1:T})$, and setting $\B{x}_t = \mathbf{I}[z_t=0] \, \B{y}^+_t + \mathbf{I}[z_t=1] \, \B{y}^-_t$.

%We extend this definition to the time series context where we would have two latent time series: $\B{y}^+$,  the latent normal time series i.e. the time series that would have been observed if there had been no anomalies in the training set; and $\B{y}^-$ the latent time series of anomalies. At each time point $t$ one of these two latent time series is observed, $\B{y}_t = \B{y}^{+}_t$ if the observed time point is normal, and $\B{y}_t = \B{y}^{-}_t$ if it is anomalous.

%for the background I would want to give quite a bit of it to Jan and Richard to write

% I think that we would want links to:
% \begin{itemize}
% 	\item time series anomaly detection roughly
% 	\item switching models
% 	\item training with label proportion
% \end{itemize}

\section{Method}

Forecasting or reconstruction models are designed to learn a model of $p^+(\cdot)$ but are typically trained directly on the observed time series $\B{x}_{1:T}$.
We propose to learn the model of  $p^+(\cdot)$  only from $\B{y}_{1:T}^+$ by inferring $z_{1:T} \sim p^z(z_{1:T})$ on the training set.
This way we can train the model only on the observed points that are normal, the ones that are equal to $\B{y}_{1:T}^+$. Depending on the model, the anomalous points can be treated as missing or the normal point can be inferred.

%In this section we describe our latent anomaly indicator model, the estimation procedure, and how the resulting model can be used to detect
%
%We present our approach for a single univariate time series, but it easily extends to a dataset of multi-variate time series where each of the time series is not only indexed by the time index but also by the dimension and the place in the dataset.

\subsection{Models}

Each of the three latent time series is modeled with a probabilistic model:
a parametrized model $p^+_\theta$ of the nominal data $\B{y}_{1:T}^+$, a fixed model $p^-$ to model the anomalous data $\B{y}_{1:T}^-$, and a model $p^z$ of the indicator time series $z_{1:T}$. %We describe each of them in turn.%$\bm{i}$, latent time series of anomaly indicator.

\paragraph{Nominal Data Model}
Many existing deep anomaly detection methods aim to model the nominal data (e.g.\ \citep{shipmon2017time, zhao2020multivariate, omnianomaly, donut, lstmvae, Zhang2019mscred, ehrlich2021spliced}), and any of them can be used to model $\B{y}^+$, the latent nominal time series. 
%In reality, the methods assume that $\B{y}^+ = \B{y}$.
%
%
Our method is agnostic to the type of model used, so that it can be combined with any probabilistic time series model, be it a deep or shallow probabilistic forecasting method, a reconstruction method, or any other type of model.
We call the model of the latent normal time series $p^+_\theta$, which is parametrised by a set of parameters $\theta$.

In our experiments we demonstrate the general setup by modeling $p^+(\B{y}^+_{1:T})$ with a simple deep probabilistic forecasting model. We decompose  $p(\B{y}^+_{1:T})$ into the telescoping product  $p(\B{y}^+_0) \prod_{t=0}^{T} p(\B{y}^+_{t+1} | \B{y}^+_{t:0})$ and, making an $l$-th order Markov assumption, approximate it with a network $p(\B{y}^+_{t+1}  |  \B{y}^+_{t:t-l}  ) = \mathcal{N}(f_\theta(\bm{y}_{t:t-l}), g_\theta(\bm{y}_{t:t-l}) )$ taking as input the last $l$ time points.

\paragraph{Anomalous Data Model}
A simple model can be used to model $p^-$, it does not need to take into account the time component as there are typically few anomalous points. 
It can be modeled with a mixture of Gaussian distributions for example, with the risk of over-fitting to the few anomalies of the train set.
We simply model $p^-$ with a uniform distribution over the domain of the training data, not assuming any prior on the kind of anomalies that we may expect.

\paragraph{Anomaly Indicator Model} 
%Most time series models can be used to model $p^z$.
We model the latent anomaly indicator with a Hidden Markov Model (HMM) with two states, state ${z}_t  = 0$ corresponds to the point being normal and state ${z}_t  = 1$ corresponds to the point being anomalous. Any kind of time series model parameterizing a Bernoulli distribution can be used to model the latent anomaly indicators, we pick an HMM as it encodes basic time dependencies while staying a simple model.

If it is available, prior knowledge about the dataset can be used to initialise the transition matrix.
The expected length of anomalous windows can be used to initialise the transition probability $p( z_{t+1} = 1 | z_t  = 1 )$.
The expected percentage of anomalous points in the dataset can be used to initialise the transition probability $p( z_{t+1} = 1 | z_t  = 0 )$.

%speak about the prior knowledge and so on

\iffalse
is the prior probability of a point being anomalous, we can either:
\begin{enumerate}
	\item set it to 0.01 by default as we assume that anomalies are rare (the one thing that we can assume in anomaly detection)
	\item use a prior that we have for this specific dataset
	\item have a prior given by labels on the dataset : e.g. 0.9 for points that are labeled as anomalous and 0.001 for the other ones
\end{enumerate}
%TODO: here we do not really explore the idea that we can use noisy labels as prior.

$p_{+}^*(\bm{y}_{t+1}) = \mathcal{N}(f_\theta(\bm{y}_{t:t-l}), g_\theta(\bm{y}_{t:t-l}) )$

We model with a forecasting model, as of now, a Gaussian parametrised by a NN : 

This model takes into account the time dependencies of the points, and so knowing which points are anomalous has two advantages:
\begin{enumerate}
	\item we do not train the model to predict them (not including them in the loss)
	\item we can replace/mask the anomalous points in the context window and so not have the model train on anomalous input
\end{enumerate}

\fi

\subsection{Training}

Our training procedure follows Monte Carlo EM \citep{wei1990monte}.
In the E-step we infer $ p^z(z_{1:T})$. % and the latent normal time series?
In the M-step we sample from $p^z(z_{1:T})$, using these samples to update $p^+_\theta$ and the transition matrix of the HMM.
Algorithm \ref{alg} sketches this procedure.

\begin{algorithm}[htpb]
	\SetAlgoLined
	\KwIn{Observed time series $\B{x}_{1:T}$, model to be trained $p^+_\theta$
		%, state space matrices $\matVar{F}$, $\matVar{H}$ given $\theta$, estimated noise variance $\matVar{R}$, state dimension $s$.
	}
	%\KwOut{$\log{p(\vecVar{f}_T|\B{t}_{1:T}, \theta)}$}
	%\tcp{Initialisation}
	%$\log{p(\vecVar{f}_T|\B{t}_{1:T}, \theta)} := 0$, $\vecVar{m}_0 \gets \matVar{0}_{m \times 1}$, $\vecVar{P}_0 \gets \matVar{0}_{m \times m}$ \\
	%	$o:= 1$ \tcp*[f]{\small Index last relevant observation} 
	\For{$e \in \{1, \dots, \text{numb\_epochs}\}$}{
		
		\tcp{E-step:}
		
		$\rightarrow$  infer $p^z(z_{1:T})$
		
		\tcp{M-step:}
		
		\For{$s \in \{1, \dots, \text{numb\_samples}\}$}{
			
			$\rightarrow$ sample indicator time series $z_s$ from $p^z(z_{1:T})$ 
			
			$\rightarrow$ perform one epoch of $p^+_\theta$ on $\B{x}_{\neg z_s}$ where the points at sampled anomalous indices are replaced
		}
		
		$\rightarrow$ update the transition matrix of the HMM
		
	}
	\caption{Monte Carlo EM for Latent Anomaly Indicator}
	\label{alg}
\end{algorithm}

\subsubsection{E-step}

We  infer $p^z(z_{1:T})$ by using the standard forward-backward algorithm for HMMs, using the following distributions:
\begin{align}
p(\B{x}_t | z_t = 0) = p^+_\theta(\B{x}_t) \\
p(\B{x}_t | z_t = 1) = p^-(\B{x}_t)
\end{align}
and $p(z_{t+1} | z_t)$ is given by the HMM transition matrix.

%(cite source, maybe Barber's book)

\iffalse
(we abuse a bit the notation here: $p(p_- | \bm{y}_t)$ is the posterior probability that the point t comes from the anomalous data distribution)

For each point, we want to infer the probability of this point coming from the anomalous model:
$$p(p_- | \bm{y}_t) = \frac{p(\bm{y}_t | p_-) p(p_-)}{p(\bm{y}_t)}  = \frac{p(\bm{y}_t | p_-) p(p_-)}{p(\bm{y}_t | p_-) + p(\bm{y}_t | p_+) } $$
First, simply using Bayes rule, then using the fact that we have:  $p(\bm{y}) = p_{+}(\bm{y}) + p_{-}(\bm{y})$ 

are obtained with our trained model for the normal data and the model for the anomalous data

\fi

\subsubsection{M-step}
%In the M-step we maximise the expectation of the probability of the observed data under $p(z)$. As this may not be analytically tractable for any model of $p^+$, especially the deep learning approaches presented above, we propose to sample from $p(\bm{i})$ and update $\theta$ using these samples.
%
%We present how we train each of the models. 

We want to train $p^+(\cdot)$ only from $\B{y}_{1:T}^+$. As most models may not allow for an analytical update using $\bm{x}_{1:T}$ and $z_{1:T}$, we propose to a Monte Carlo approximation of the expectation under $p^z(z_{1:T})$.
We draw multiple samples from $p^z(z_{1:T})$ giving us possible normal points on which $p^+_\theta$ can be trained.
Each path sampled gives us a set of observed points that can be considered as coming from the normal data distribution $p^+$.
We maximise the probability of these points under $p^+_\theta$, treating the points coming from $p^-$ points as missing.

Depending on the choice of model for $p^+_\theta$, one may not be able to simply ignore anomalous points and they would have to be imputed. For deep forecasting or reconstruction models for example the model has to be given an input for each time point.
In these cases, we propose to impute the point with the forecast or reconstruction obtained from $p^+_\theta$ at the last M-step.
This way, we use $p^+_\theta$ to infer the time points of $ \B{y}_{1:T}^+$ that were not observed.
With this method we can recover the full $\B{y}_{1:T}^+$ time series and train $p^+_\theta$ on it.
%For reconstruction models, if an anomalous point is to be reconstructed, the loss for this point can simply be ignored.

% replacing the anomalous points
% not computing the loss at the anomalous outputs
%(could cite Max Wellings' paper on propagating the uncertainty through the network)

%\paragraph{Model of the anomalous data}
Depending on the choice of model for $p^-$, one can update it using the points that are sampled as coming from $ \B{y}_{1:T}^-$. 
%With out choice of using the uniform distribution, we do not have
%If we go with the uniform distribution, there is no need for update.

%\paragraph{HMM} 
We can update the transition matrix of the HMM with the classical M-step. The average number of transitions from one state to the next in the samples from $p^z(z_{1:T})$ become the new transition probabilities.

\iffalse
We train the forecasting model by masking the inferred anomalous points from the input vector.

We sample independently for each time point if it comes from or not. If we sample it, we replace it by the predicted mean $ f_\theta(\bm{y}_{t:t-l})$ obtained at the last epoch.

In addition we ignore the loss for the points that we sample as coming from the anomalous distribution.

This procedure allows to use mini batch training very easily:

\begin{enumerate}
	\item we sample the points to ignore for the whole training set
	\item we replace them in the input time series
	\item we can do minibatches using this modified input time series
\end{enumerate}

For each training epoch, we sample 20 time from the distribution.
\fi

\subsection{Inference}

At inference time, we propose to use the HMM to perform filtering on $z$ and infer if incoming points are more likely to be drawn from $p^+$ or $p^-$.
If an incoming point $\B{x}_t$ is more likely to be coming from $p^-$ it can be treated as missing or replaced with a sample from $p^+_{\theta t}$ or by its mode.
This way we ensure that the trained model is only used on points coming from $\B{y}_{1:T}^+$.

%Once it is trained, the model $\mathcal{M}^+$ can be deployed.
%
%We propose to not deploy the model alone but rather to still use the HMM to do filtering as described above and to ignore or replace the points that are more likely to come from the anomalous distribution than from the normal distribution.
%In the classical setting where one obtains one point at a time and has to flag it as normal or anomalous, the HMM can o
%
%Depending on the requirements, we would use the filtering of the identifier to infer if an incoming point is anomalous or not, if it is it will be treated as missing.

\section{Experiments}
\iffalse

\fi

%-> a pass through the paper
%
%-> cite Spliced Binned Pareto work

We make our code available with an illustration notebook.
\footnote{ \url{https://github.com/Francois-Aubet/gluon-ts/blob/monte_carlo_em_masking_ notebook/src/gluonts/nursery/anomaly_detection/Monte-Carlo-EM- for-Time-Series-Anomaly-Detection-demo-notebook.ipynb}}

\paragraph{Model}
%We use two different models which we train normally and with our method.
%
We evaluate our approach with a simple forecasting model on both anomaly detection and forecasting tasks. We show the performance of the model when trained in a standard way and when trained with our procedure, which we call our procedure Latent Anomaly Indicator (LAI).
%In each of the following scenarios, we use a very simple forecasting model to show the results that our method can bring when used on top of it.
We use a simple Multi-Layer Perceptron (MLP) model to parametrise the mean and the variance of a predictive Gaussian distribution. It takes as input the last 25 points.

\paragraph{Datasets}
For the anomaly detection evaluation, we use the \textbf{Yahoo} dataset,  published by Yahoo labs.\footnote{\url{https://webscope.sandbox.yahoo.com/catalog.php?datatype=s&did=70}}
It consists of 367 real and synthetic time series, divided into four subsets (A1-A4) with varying level of difficulty.
The length of the series vary from 700 to 1700 observations.
Labels are available for all the series.
We use the last 50\% of the time points of each of the time series as test set, like \citep{ren2019time} did,  and split the rest in 40\% training and 10\% validation set.
We evaluate the performance of the model using the adjusted F1 score proposed by \citet{donut} and subsequently used in other work.

In addition, we evaluate the method on forecasting tasks using the commonly used \textbf{electricity} dataset \cite{electricity}, composed of 370 time series of 133k points each. Given the length of the dataset, we sub-sample it by a factor 10. We select the last 50\% of the points of each time series for testing.
We scale each time series using the median and inter-quartile range on the train~set.
%We corrupt it by adding point outliers in the training set, to see if our method allows to train the model despite the outliers.

\subsection{Visualization on synthetic data}

\begin{figure}[t]
	\begin{subfigure}{.5\textwidth}
		% include second image
		\includegraphics[width=.99\textwidth]{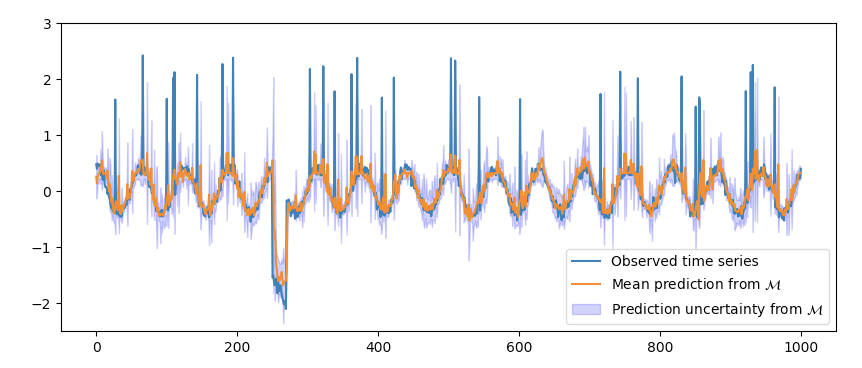}  
		\caption{The fit of a simple MLP without LAI.}
		\label{fig:normal_model_fit}
	\end{subfigure}
	\begin{subfigure}{.5\textwidth}
		% include first image
		\centering
		\includegraphics[width=.99\textwidth]{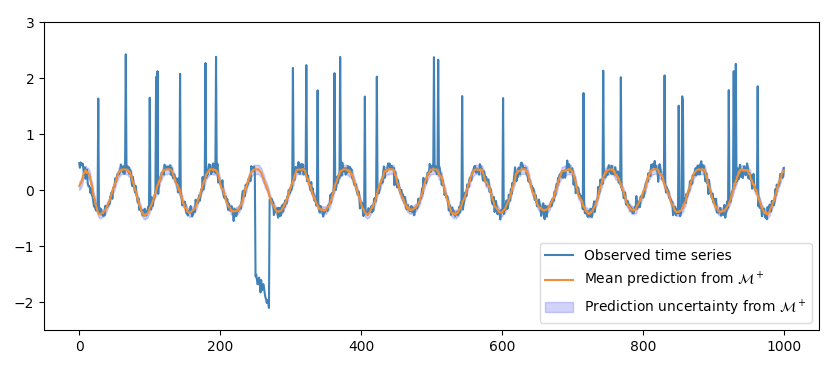}  
		\caption{The fit of a simple MLP with LAI.}
		\label{fig:masking_model_fit}
	\end{subfigure}
	\begin{subfigure}{.5\textwidth}
		% include second image
		\includegraphics[width=.99\textwidth]{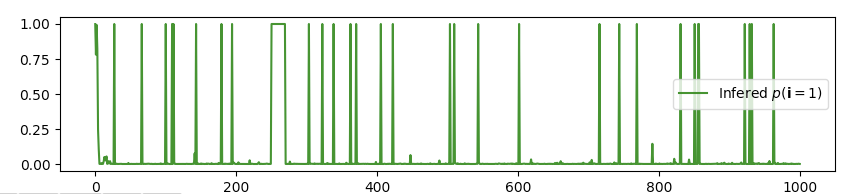}  
		\caption{The time series of latent anomaly indicator $p(z_t = 1)$.}
		\label{fig:masking_pi}
	\end{subfigure}
	\caption{ We fit a MLP on this simple synthetic time series with anomalies. (a) shows the fit of the model trained in a conventional way, (b) shows the fit of the model trained as we propose to, (c) show the inferred $p(z_{1:T})$ distribution at the end of the training.
	}
\label{fig:synthetic_ts}
\end{figure}

Figure \ref{fig:synthetic_ts} visualizes the advantage of the method on a simple sinusoidal time series with the simple MLP for $p^+_\theta$. We generate a synthetic time series and inject outliers in it. 
%We use a simple feed forward neural network taking as input the last 25 time points to parametrise a Gaussian distribution on the next time point.
%
We observe that our approach allows to train the model $p^+_\theta$ while ignoring the outliers in the data, %we can see that our training procedure allows to recover the latent normal data, 
whereas the outliers heavily influence the model trained conventionally.
We observe from figure \ref{fig:masking_pi} that the model is able to infer accurately which of the training points are likely to be anomalous.

\subsection{Time series anomaly detection}

Table \ref{table:test_set} shows the F1 score of the model with and without LAI on the different subsets of the Yahoo dataset.
We train one MLP  on each of the time series and average the F1 scores obtained on the different time series of the subset.
We observe that using our approach greatly improves the performance of the model.
%
%We note that the performance is worse on A1 which is explained by the fact that this 

\begin{table}[h]
	\caption{F1 score on the different subsets of the Yahoo dataset.}
	\label{table:test_set}
	\centering
	\begin{tabular}{l|cccccccccccc}
		\toprule
		Model & A1  & A2  & A3  & A4  \\ 
		\midrule
		MLP & 33.64  & 53.28 & 63.25  & 47.30  \\
		MLP + LAI  & 41.84 & 87.26 & 87.91 & 61.62    \\
%		\midrule
%		LSTM & ? & ?  & ? & ?   \\
%		LSTM + LAI  & ? & ?  &  ? & ?    \\
		\bottomrule
	\end{tabular}
\end{table}

\iffalse
ARIMA:
Yahoo A1:
Normal average:  0.3363546742021299
LAI average:  0.4183863827943147

Yahoo A2:
Normal average:  0.5328321386575404
LAI average:  0.8726312951525773

Yahoo A3:
Normal average:  0.6324923474686052
LAI average:  0.8790611366231169

Yahoo A4:
Normal average:  0.4730399846547243
LAI average:  0.6161590733402865

LSTM:

\fi

In addition to the improved F1 score, we compare the inferred anomalous points on the training set with the actual labeled anomalous points.
Table \ref{table:train_set} shows the F1 score on the training set when using the anomaly indicator as anomaly score.
We observe that our method allows to find accurately the anomalies present in the training set.
%, and  allows to label a training set for supervised anomaly detection methods.
While the training and test sets are different, we propose that the higher F1 on the train set is due to the fact that the model can use the whole training set to infer if a point is anomalous, and not only the past points.

\begin{table}[H]
	\caption{F1 score on the training set the different subsets of the Yahoo dataset using the inferred $p(z_{t}=1)$ as anomaly score.}
	\label{table:train_set}
	\centering
	\begin{tabular}{l|cccccccccccc}
		\toprule
		Model & A1  & A2  & A3  & A4  \\ 
		\midrule
%		MLP & ?  & ? & ?  & ?  \\
		MLP + LAI  & 59.48 & 94.02 & 81.89 & 73.77    \\
%		LSTM & ? & ?  & ? & ?   \\
%		LSTM + LAI  & ? & ?  &  ? & ?    \\
		\bottomrule
	\end{tabular}
\end{table}

\subsection{Forecasting using a corrupted train set}

Our method can be used more generally to train a forecasting model on a forecasting dataset containing anomalies. 
We take the electricity forecasting dataset and inject point outliers in the training set so that about 0.4\% of the training point have an added or subtracted spike.
Table \ref{table:forecasting} shows the mean absolute error (MAE) on the test set in the setting where the original train set is used and in the setting where the noisy train set is used.
We see that using our method allows to reduce significantly the increase in error from the outliers in the training set, only 0.0146 increase in the mean absolute error versus 0.0542 when training the model normally.

\begin{table}[H]
	\caption{MAE on electricity with and without injecting point outliers in the train set}
	\label{table:forecasting}
	\centering
	\begin{tabular}{l|cccccccccccc}
		\toprule
		Model & electricity & electricity + outliers \\ 
		\midrule
		MLP & 0.1551 & 0.2092   \\
		MLP + LAI  & 0.1558 &  0.1704    \\
		\bottomrule
	\end{tabular}
\end{table}

%\subsection{Time series anomaly detection with a forecasting model (supervised)}
%\subsection{Same method but other kind of model?}

\section{Conclusion}

We present LAI, a method that  can be used to wrap any probabilistic time series model to perform anomaly detection without being impacted by unlabeled anomalies in the training set.
We present the details of the approach and propose preliminary empirical results on commonly used public benchmark datasets.
The approach seems to greatly help both for anomaly detection tasks and for training a forecasting model on a contaminated training set.

One can extend this work by wrapping other bigger models such as OmniAnomaly \citep{omnianomaly} or state-of-the-art forecasting models \citep{benidis2020neural}.
Finally, with our current method at inference time, one has to decide at each incoming point if it is to be replaced or not, one could use particles which would mimic the Monte Carlo approach of the training time.

\newpage
\
\newpage
\bibliography{references}
\bibliographystyle{icml2021}

%\section*{Acknowledgements}
%
%\textbf{Do not} include acknowledgements in the initial version of
%the paper submitted for blind review.

\nocite{langley00}

%\bibliography{example_paper}
%\bibliographystyle{icml2021}

\iffalse
\newpage
\appendix
%\section{Do \emph{not} have an appendix here}

\section{Nomenclature}

Here we define all the nomenclature:

we need:
\begin{itemize}
	\item $\B{y}$ observed time series
	\item $\B{y}_t$ observed time series at time point $t$
	\item $\B{y}^+$ latent normal time series
	\item $\B{y}^-$ latent anomalous time series
	\item $\bm{i}$ latent time series of anomaly indicator
\end{itemize}

\fi

%%%%%%%%%%%%%%%%%%%%%%%%%%%%%%%%%%%%%%%%%%%%%%%%%%%%%%%%%%%%%%%%%%%%%%%%%%%%%%%
%%%%%%%%%%%%%%%%%%%%%%%%%%%%%%%%%%%%%%%%%%%%%%%%%%%%%%%%%%%%%%%%%%%%%%%%%%%%%%%

\end{document}